\documentclass[letterpaper, 10 pt, conference]{ieeeconf}  

\IEEEoverridecommandlockouts                              

\usepackage{graphicx}
\usepackage{amsmath}
\usepackage{amssymb}
\usepackage{algorithm}
\usepackage{algpseudocode}
\usepackage[dvipsnames]{xcolor}
\usepackage[labelformat=simple]{subcaption}
\usepackage[bookmarks=true,colorlinks=true]{hyperref}
\usepackage[normalem]{ulem}
\useunder{\uline}{\ul}{}

\title{\LARGE \bf
WayFAST: Navigation with Predictive Traversability in the Field 
}

\author{Mateus V. Gasparino$^{1}$, Arun N. Sivakumar$^{1}$, Yixiao Liu$^{1}$, Andres E. B. Velasquez$^{1}$, \\
Vitor A. H. Higuti$^{2}$, John Rogers$^{3}$, Huy Tran$^{4}$, and Girish Chowdhary$^{1}$
\thanks{Project code repository: \url{https://github.com/matval/WayFAST}}%
\thanks{$^{1}$Field Robotics Engineering and Science Hub (FRESH), Illinois Autonomous Farm,
        University of Illinois at Urbana-Champaign (UIUC), IL}%
\thanks{$^{2}$EarthSense Inc., Champaign, IL, USA}%
\thanks{$^{3}$DEVCOM Army Research Lab, Adelphi, MD, USA}%
\thanks{$^{4}$Dept. of Aerospace Engineering,
       UIUC, IL, USA}%
\thanks{{Correspondence to \tt\small \{mvalve2,girishc\}@illinois.edu}}%
\thanks{{This work was supported in part by ARL \#W911NF2020184, NSF STTR \#1951250, NSF NRI 2.0 NIFA \#2021-67021-33449, and AIFARMS \#1024178.}}%
}

\begin{document}

\maketitle
\thispagestyle{empty}
\pagestyle{empty}

\begin{abstract}

We present a self-supervised approach for learning to predict traversable paths for wheeled mobile robots that require good traction to navigate. Our algorithm, termed WayFAST (Waypoint Free Autonomous Systems for Traversability), uses RGB and depth data, along with navigation experience, to autonomously generate traversable paths in outdoor unstructured environments. Our key inspiration is that traction can be estimated for rolling robots using kinodynamic models. Using traction estimates provided by an online receding horizon estimator, we are able to train a traversability prediction neural network in a self-supervised manner, without requiring heuristics utilized by previous methods. We demonstrate the effectiveness of WayFAST through extensive field testing in varying environments, ranging from sandy dry beaches to forest canopies and snow covered grass fields. Our results clearly demonstrate that WayFAST can learn to avoid geometric obstacles as well as untraversable terrain, such as snow, which would be difficult to avoid with sensors that provide only geometric data, such as LiDAR. Furthermore, we show that our training pipeline based on online traction estimates is more data-efficient than other heuristic-based methods.

\end{abstract}

\section{Introduction}

Waypoint-based path programming methods for field robots are slow, ineffective, and cumbersome in outdoor and unstructured environments. This has become a major barrier for deployment of field robots in wooded, obstacle prone, or poorly mapped areas, such as remote outposts, contested areas, forests, and agricultural fields. The user has to plan the path around all possible obstacles, including rocks, trees, holes, etc. which is impractical. 

As such, algorithms for traversability determination have remained an active area of research (see Section II Related Work). The main challenge here is that heuristics-based methods are difficult to design and often lead to false positives in outdoor environments where geometric traversability is not always equivalent to actual traversability. For example, tall grass could be detected by LiDAR or tactile sensors as geometric obstacles, when in fact, these are obstacles that the robot could  traverse through. Moreover, in addition to predicting obstacle free paths, the traversability system should also predict expected traction coefficient to ensure successful autonomous navigation \cite{ding20222,kayacan2018embedded}.

\begin{figure}[t]
    \centering
    \includegraphics[trim={0cm 3cm 0 7cm}, clip, width=0.96\linewidth]{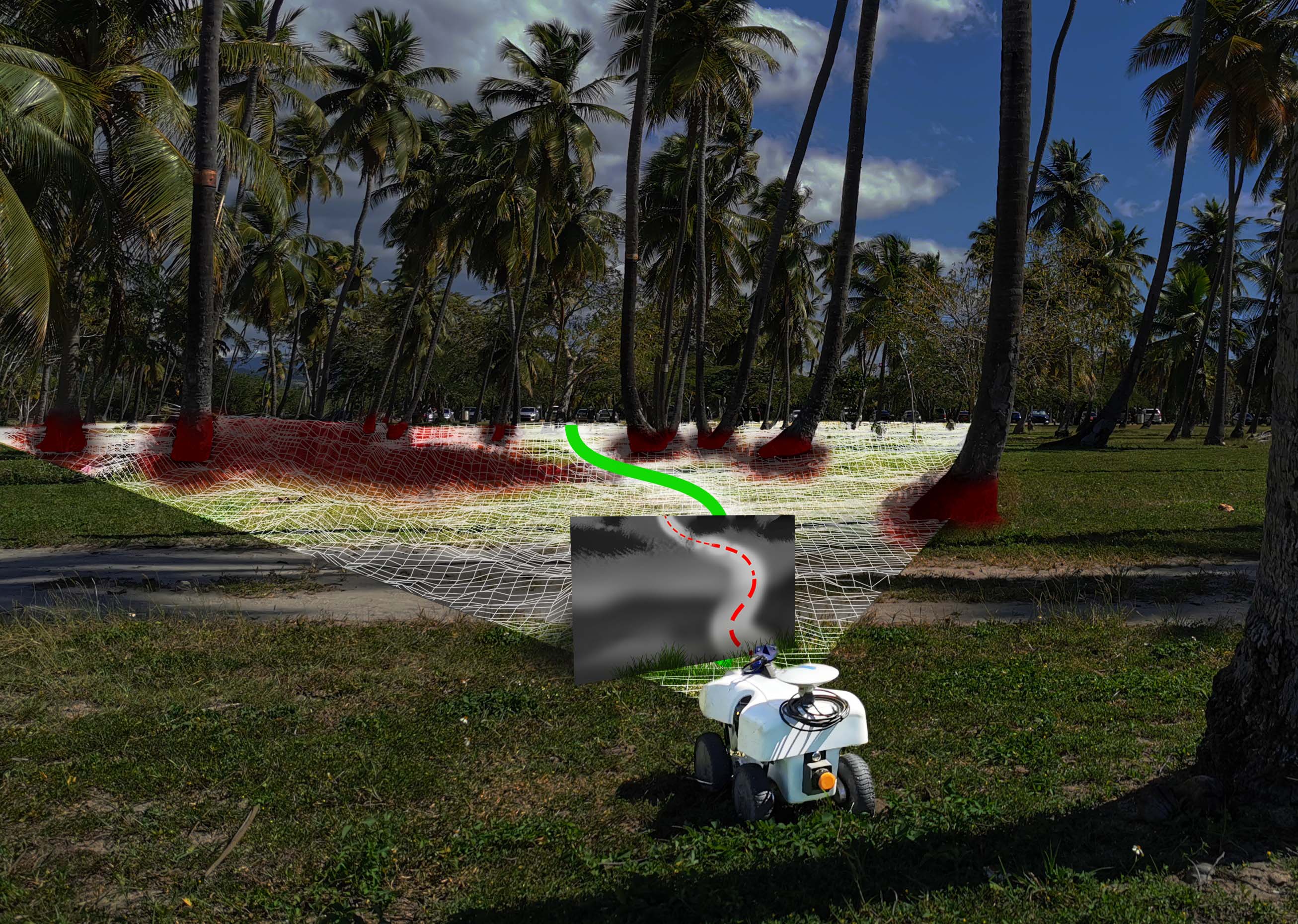}
    \caption{We present WayFAST, a \textbf{Way}point \textbf{F}ree \textbf{A}utonomous \textbf{S}ystem for \textbf{T}raversability prediction that learns to predict obstacle free paths of good traction outdoors.}
    \label{fig:wayfast}
    \vspace{-0.1in}
\end{figure}

Learning-based methods to determine traversability based on navigation experiences could lead to more robust navigation by avoiding heuristics \cite{howard2006towards,kahn2021badgr,kim2006traversability,wellhausen2020safe,miki2022learning}. However, learning traversability and robot kinodynamic models together, as done in \cite{kahn2021badgr,loquercio2018dronet,chiang2019learning}, may lead to oscillatory paths. Furthermore, coupling the problem of learning traversability and robot model is unnecessary because kinodynamic models for wheeled robots are well known, as is evident by success of Model Predictive Control (MPC) \cite{kayacan2018embedded,lindqvist2020nonlinear,sun2019tracking} or Model Predictive Path Integral (MPPI) control \cite{williams2017model,williams2016aggressive,gandhi2021robust}. Existing methods have also used heuristics such as collisions, robot attitude, or bumpiness to determine traversability labels. However this approach is limiting, for example, many untraversable terrains, such as snow, sand, or mud are not bumpy. Finally, current methods only provide a binary mask on traversable or non-traversable regions; however, reality is far more complicated. For example, some regions may be \textit{seem} traversable and even lead to a shorter path, but not be preferred due to potential lack of traction, such as a sandy patch or an icy path.

\subsection{Contributions}

In this paper, we present a new modular approach to speed-up field robot path programming that learns to drive the robot on paths of optimal traction. We term our method \textbf{WayFAST}: \textbf{Way}point \textbf{F}ree \textbf{A}utonomous \textbf{S}ystems for \textbf{T}raversability, and demonstrate that it significantly reduces field robot path programming time and effort by letting the user simply pick the target point where the robot needs to travel, illustrated in Fig. \ref{fig:wayfast}. Our method is completely self-supervised and heuristic free; that is, it does not need any manual labels or heuristics to be defined for learning traversable regions. Our method is modular, being able to plug into proven state estimators, such as Extended Kalman Filters or Moving Horizon Estimators \cite{thrun2002probabilistic,liu2016moving}, and model predictive controllers, such as traction adaptive Nonlinear Model Predictive Control or MPPI control \cite{kayacan2018embedded,williams2017model,williams2016aggressive,gandhi2021robust}. We show through extensive experimentation in various challenging outdoor environments in two entirely different geographies (US Midwest and Puerto Rico) that our method outperforms existing methods and leads to oscillation free autonomous navigation.

\section{Related Work}

\textbf{Classical navigation.} Global Navigation Satellite Systems (GNSS) based waypoints are the standard method for robot path programming, but this is cumbersome in unstructured outdoor environments. Classical approaches either do reactive obstacle avoidance through heuristics based on sensor data or simultaneously build a map of the environment and determine the robot's position in the map (SLAM) to navigate \cite{thrun2002probabilistic,siegwart2011introduction}. Both approaches only perceive the environment geometrically and assume all geometric obstacles are untraversable but ignore valuable semantic information (e.g., tall grass is geometrically untraversable but a robot can likely go through it). In addition, SLAM methods are computationally expensive, require the environments to have textural variations to reliably track features, and are sensitive to visual aliasing in loop-closure. This limits their performance in challenging unstructured outdoor environments (like forests). In addition, it is not necessary to construct a global map if a navigation system can perceive both geometry and semantics in its local environment.

\textbf{Learned navigation in indoor environments.} Learning-based approaches have been used in indoor robotics to tackle some of the above challenges but they do not generalize to outdoor environments. \cite{bowman2017probabilistic, mccormac2017semanticfusion} fused semantic information to classical sparse and dense SLAM systems but the other limitations in outdoor environments still exist. \cite{gupta2017cognitive,bansal2020combining} use privileged information during training to learn navigational affordances in a modular manner but assume access to perfect odometry. \cite{wijmans2019dd,zhu2017target} use end-to-end learning to output control commands from input images by training a reinforcement learning (RL) agent. However, RL methods are generally sample inefficient and hence these methods require simulations from photorealistic scans for training.

\textbf{Learned navigation in structured outdoor environments.}
Autonomous driving is a common example of a structured outdoor environment. Different modular approaches that use the structure of lanes to learn affordances have been proposed \cite{chang2018deepvp,chen2015deepdriving}.
In agricultural settings, \cite{sivakumar2021learned} uses crop rows as a reference to train a relative state estimation network to autonomously drive a robot in the lane between two crop rows. However, wooded areas, remote outposts, or other rural landscape does not always have reliable repeating structure.

\textbf{Supervised and self-supervised learning in unstructured outdoor environments.} Learning traversability is an emerging area of research.
\cite{valada2017adapnet} takes a supervised learning approach, where a network model is trained for semantic segmentation and shown to be effective for autonomous navigation. However, this requires manual labeling, which is tedious.
For quadrupeds, \cite{wellhausen2019should} presents a method to generate semantic segmentation and dense image predictions based on force sensors. The semantic segmentation is created with weak supervision and the authors show it is possible to use the neural network predictions to choose paths that require less force for quadruped robots locomotion. These tactile force sensors are not common for wheeled robots, add cost and complexity, and although they can be used for ground preference, they are not useful for obstacles. \cite{wellhausen2020safe} uses anomaly detection with normalizing flow to detect out of distribution situations from a navigation training dataset. This produces a binary map of potentially safe areas for navigation. However, binary estimation oversimplifies the perception, making partially traversable paths equally important to fully traversable ones. BADGR \cite{kahn2021badgr} trains an end-to-end model-based RL policy with labels automatically created from embedded sensors. Using different sensors and a set of rules to define collisions, BADGR can predict collision probability and uses a sampling based algorithm to avoid collision and bumpy terrains. BADGR's self created dataset avoids explicit human labeling, but still requires heuristics to generate collision labels. We instead leverage known kinodynamic models to create a traversability predictor that can avoid obstacles and predict paths of good traction.

\section{System Design}

WayFAST is a modular navigation system based on traversability estimation for unstructured outdoor environments. RGB and depth images from on-board camera on the robot are processed through a convolutional neural network to predict traversability. The generated traversability map along with a defined kinodynamic model is then used for model predictive control. Given a goal, the system is able to minimize a cost function that safely guides the robot to the target location. In this section, we describe the robotic platform and various modules of WayFAST in detail.

\begin{figure*}[htp]
    \centering
    \includegraphics[trim={0cm 3cm 0 6.5cm}, clip, width=0.76\linewidth]{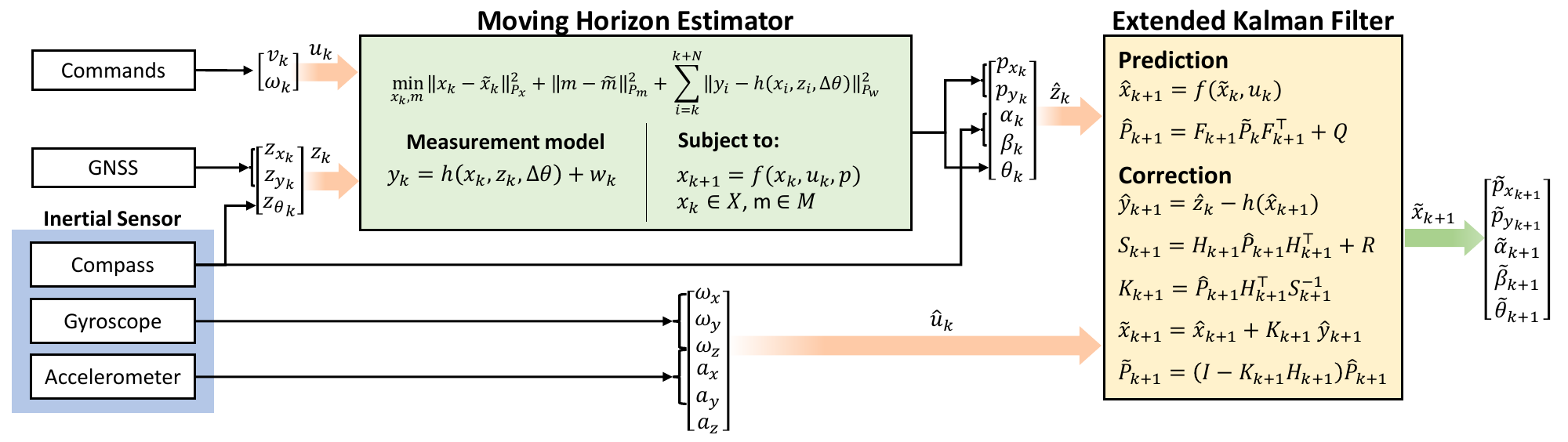}
    \caption{Our state estimation scheme, where the green box shows the Nonlinear Moving Horizon Estimator that estimates two-dimensional robot's states and unknown model parameters. The yellow box shows the Extended Kalman Filter used to augment the estimates to a 2.5-dimensional space and integrate the moving horizon estimations with a prediction model.} 
    \label{fig:estimator}
    \vspace{-0.1in}
\end{figure*}

\subsection{Robot platform description}

TerraSentia (by Earthsense Inc.) is a skid-steer robot, which is a type of robot commonly used in many applications due to its mechanical simplicity, such as agricultural fields, industrial automation, mining, and planetary exploration \cite{higuti2019under,liao2018model,huh2011development,reid2015artemis}. We used an Intel RealSense D435i camera on top and this provides color and depth images (with usable range of 5 meters, which is sufficient as the sensor is pointed towards the terrain). We used a Nvidia Jetson Xavier NX computer in the robot to run our algorithms.

Affine kinodynamic model, given by \eqref{eq:kino-model}, describes skid-steer robots, like
TerraSentia. We use this model to formulate the estimator that provides self-supervised labels for training and also in the model based controller.

\begin{equation}
    \dot{x}(t)
    =
    \begin{bmatrix}
        \dot{p}_x(t) \\
        \dot{p}_y(t) \\
        \dot{\theta}(t)
    \end{bmatrix}
    =
    \begin{bmatrix}
        \mu \cdot cos(\theta(t)) & 0 \\
        \mu \cdot sin(\theta(t)) & 0  \\
        0 & \nu
    \end{bmatrix}
    \begin{bmatrix}
        v(t) \\
        \omega(t)
    \end{bmatrix}
    \label{eq:kino-model}
\end{equation}

$x(t)$ is the state vector composed of  $p_x(t)$, $p_y(t)$ and $\theta(t)$ representing position in x and y axis and heading angle, respectively, in the world coordinate frame. $\mu$ and $\nu$ are unknown parameters and can be related to a skid coefficient caused by the interaction between the wheels and terrain. $v(t)$ and $\omega(t)$ are the commanded linear and angular velocities, which are the control inputs.

\subsection{Generating Traversability Labels with NMHE} \label{subsection:mhe}

We use a nonlinear moving horizon estimator (NMHE) to generate traversability labels to train our prediction network. The NMHE uses the system model shown in \eqref{eq:kino-model} and synchronously runs with GNSS on the robot. We define $\mu$ and $\nu$ shown in \eqref{eq:kino-model} as the traversability coefficient. To estimate this coefficient, we use the commanded action as system input. This allows us to estimate how the robot behaves when action command is applied. Traversibility coefficients should be one when the robot perfectly moves according to the commanded action and zero when the robot is stuck at a place such as when it encounters an obstacle or untraversable terrain.

NMHE is ideal for instantaneous traversability estimation since it deals with constrained states and parameters. We assume the measurements model $h(\cdot)$ is subject to an additive measurement noise $w(t)$ such that the measurements $y(t)$ may be defined as
\begin{equation}
    y(t)=h(x(t),z(t),\Delta\theta)+w(t).
\end{equation}

We solve the following finite horizon optimization formulation to obtain the system states and unknown parameters $\mu$, $\nu$, and $\Delta\theta$
\begin{align} \label{eq:mhe-optimization}
    \begin{gathered}
        \min_{x_{k:k+N}, \mu, \nu, \Delta\theta} ||x_k-\Tilde{x}_k||^2_{P_x} + 
        ||m-\Tilde{m}||^2_{P_m} + \\
        \sum_{i=k}^{k+N} ||y_i-h(x_i,z_i,\Delta\theta)||^2_{P_w}
    \end{gathered}
\end{align}
subject to the constraints
\begin{align} \label{eq:mhe-constraints}
    \begin{gathered}
        x_{k+1} = f(x_k,u_k) \\
        \mu, \nu \in [0,1] \\
        \Delta\theta \in [-\pi,\pi)
    \end{gathered}
\end{align}

$\Tilde{x}_k$ is the initial estimated state vector, $\Delta\theta$ is the offset between true North and compass estimated North, $m=[\mu, \nu, \Delta\theta]^\top$ is the vector of parameters, $\Tilde{m}$ is the estimated vector of parameters from the previous iteration, and $N \in \mathbb{N}$ is the length of the horizon. Only the system states and unknown parameters $\mu$, $\nu$, and $\Delta\theta$ are estimated in the optimization. The measurement equation model, subject to additive noise, is defined as
    \begin{equation*}
        w_k = y_k-h(x_k,z_k,\Delta\theta) =
        \begin{bmatrix}
            p_{x_k} - z_{x_k} \\
            p_{y_k} - z_{y_k} \\
            \theta_k - (z_{\theta_k} + \Delta\theta)
        \end{bmatrix}
    \end{equation*}
where $z_k = [z_{x_k}, z_{y_k}, z_{\theta_k}]^\top$, $z_{x_k}$ and $z_{y_k}$ are the pair of measured position coordinates from GNSS, and $z_{\theta_k}$ is the measured heading angle from the embedded inertial sensor. For estimation purposes, we assume the parameters are constant in a short horizon $N$.

The estimator is followed by an Extended Kalman Filter (EKF) \cite{thrun2002probabilistic, farrell2008aided} which uses a prediction and a measurement model to estimate the robot's states in a 2.5-dimensional space. In this implementation, we disregard the state associated with the height and assume a planar navigation with three-dimensional rotation. Our prediction model is synchronized with the embedded inertial sensor which fills the predictions in a higher frequency while NMHE deals with constraints. Figure \ref{fig:estimator} shows how the NMHE data is used in the EKF where NMHE outputs are used as measurements to correct the EKF predictions. The gyroscope outputs the angular velocities $\omega_x$, $\omega_y$, $\omega_z$, and the accelerometer sensor outputs the linear accelerations $a_x$, $a_y$, $a_z$,  while $\hat{u}_k$ is the composition of these six measurements. The inertial compass outputs the angle measurements $z_{\theta_k}$, $\alpha_{k}$, $\beta_{k}$, associated with the robot's yaw, pitch, and roll angles, respectively. The vector $\hat{z}_k$ is composed of the NMHE outputs $p_{x_{k}}$, $p_{y_{k}}$, $\theta_{k}$ and the angles $\alpha_{k}$ and $\beta_{k}$.

\subsection{Self-Supervised Learned Perception for Traversability}

\begin{figure}[htp]
    \centering
    \includegraphics[trim={0cm 8cm 0cm 8cm}, clip, width=0.98\linewidth]{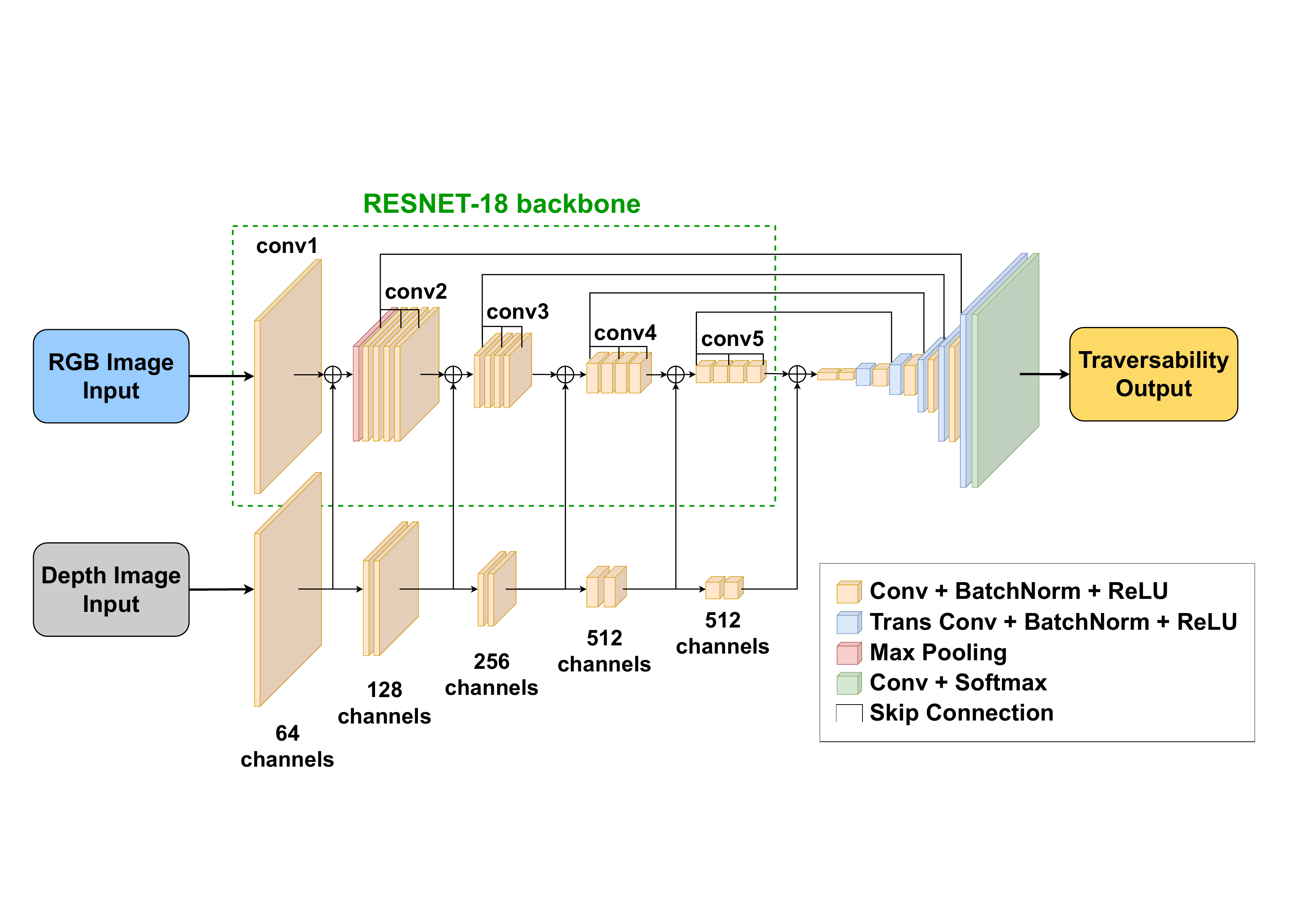}
    \caption{Our TravNet architecture, which fuses RGB and depth information to predict traversability in the input image.} 
    \label{fig:net-architecture}
    \vspace{-0.1in}
\end{figure}

We present a convolutional neural network named TravNet to estimate traversability coefficient values in the image space. Traversability coefficient represents how much control effort is needed to reach a defined location. Traversability prediction in image space provides traversability coefficient for trajectories within robot's field-of-view which is then used to plan the locally optimal path to the goal. TravNet takes in RGB and depth images of size 424×240 and outputs a single channel traversability prediction image of same size as input. TravNet uses a ResNet-18 backbone \cite{he2016deep} pretrained on ImageNet \cite{deng2009imagenet}. We truncate the ResNet-18 network before the average-pooling layer, and add a bottleneck block with two convolutional layers followed by a decoder with four blocks of convolutional layers. A second branch encodes the depth information and fuses it with the ResNet-18-based RGB encoder after each convolutional block \cite{jiang2018rednet, sun2020real}. Figure \ref{fig:net-architecture} shows TravNet's architecture and how depth information is fused with RGB to predict traversability.

\begin{figure}[htp]
    \centering
    \includegraphics[trim={0cm 0cm 0 0cm}, clip, width=0.96\linewidth]{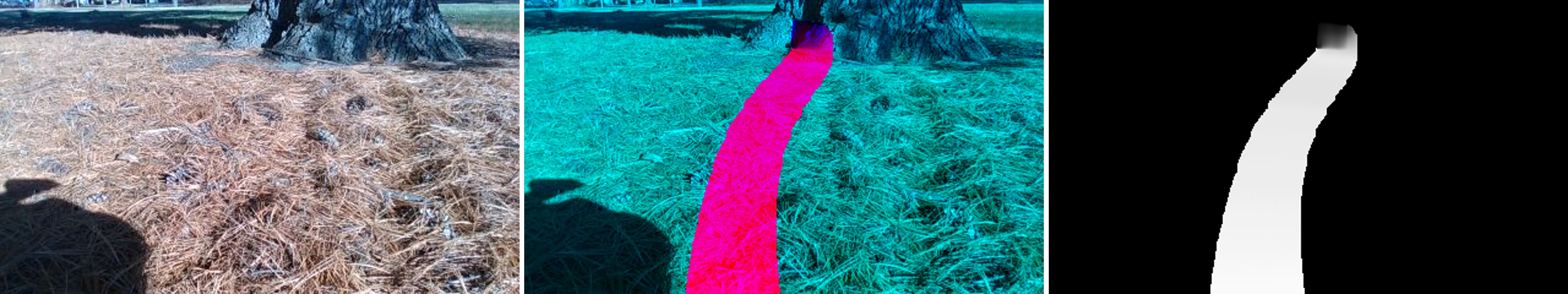}
    \caption{An example of labeled data created by projecting traversability estimations onto an image.}
    \label{fig:label-image}
    \vspace{-0.1in}
\end{figure}

\begin{figure}[htpb]
    \centering
    \includegraphics[trim={0cm 3cm 0cm 3cm}, clip, width=0.95\linewidth]{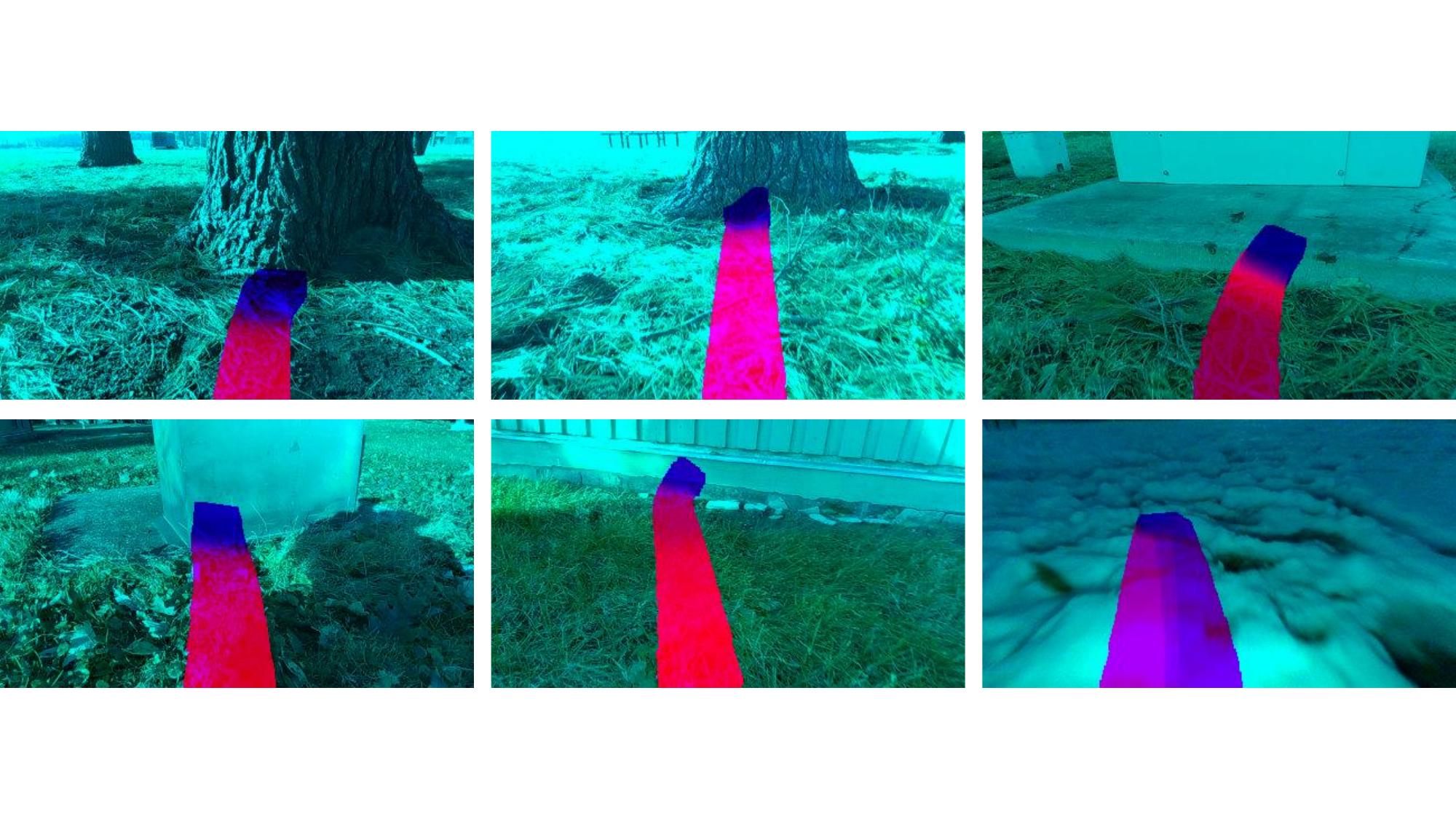}
    \caption{Sample traversability estimations projected onto input images. Darker areas represent lower estimated traversability, while brighter areas show higher estimated traversability.} 
    \label{fig:dataset}
    \vspace{-0.1in}
\end{figure}

To acquire data for training TravNet, we manually drove the robot on forest-like and semi-urban outdoor environments. We collected camera observations and state estimations using our estimator for a period of about 2 hours, which is equivalent to a dataset of 15000 images. Figure \ref{fig:dataset} shows examples of untraversable areas encountered in the dataset. The projected path shows the robot's future trajectory for that given time instant, and the color intensity shows the traversability coefficient estimation. 

To train TravNet, we follow a similar approach to \cite{wellhausen2019should}, except that we use our online state estimation with NMHE + EKF to generate the training labels. During data collection, the estimator is used to record the robot's position and traversability coefficients. To prepare the training and validation data, recorded image stream is downsampled to 2 fps and the traversed path is projected onto the image space using known transformation matrices to create the label image for each corresponding RGB image. An example is shown in Fig. \ref{fig:label-image}. Note the area on the tree is darker, and therefore, the collision point is also included as a label in the dataset.

The collected data is highly imbalanced with more examples of successful navigation of the robot (traversability coefficient close to one) compared to failures when the robot gets stuck (traversability coefficient close to zero). To address this issue, we use a label distribution smoothing technique to more heavily weight data from failures, based on the approach shown in \cite{yang2021delving}. To implement this method, we divide the traversability coefficients into five non-overlapping intervals and we calculate the histogram of the traversability distribution for the entire dataset, where $P_i$ is the probability value for the $i^{th}$ interval. The weight for each traversability interval can be calculated as $w_i = (1 - P_i)/(n_{bins}-1)$, where $n_{bins}$ is the number of defined intervals. When calculating the loss to train TravNet, each traversability value in the image is multiplied by this calculated weight depending on the range it belongs to.

To account for sparse labels, we create a $mask$ binary image. This mask contains ones where labels are present (where the robot navigated) and zeros everywhere else. This mask is multiplied by both the $label$ image (with traversability values projected where the robot navigated) and the predicted image $pred$ (from the network model). The final training loss is defined as $\mathcal{L} = \sum_{i=1}^N w_i |mask_i \cdot pred_i \text{ - } mask_i \cdot label_i | / N$, where $N$ is the total number of pixels.

\subsection{Traversability-based Model Predictive Controller}

To drive the robot on a safe path to reach the goal location, we formulated a nonlinear model predictive controller (NMPC) that leverages TravNet's traversability prediction. As described in section \ref{subsection:mhe}, we use the kinodynamic model in \eqref{eq:kino-model} to predict the robot's states. In this case, $\mu$ is predicted as a function of the robot's position, while the parameter $\nu$ is held constant. We predict $\mu$ using TravNet and set $\nu$ as the average value from estimations in the dataset. The TravNet output is used as lookup table, and bilinear interpolation is used to interpolate pixel values into a continuous function. We define this continuous function as $\mu(x,y)$, parameterized as a function of the system states $p_x$ and $p_y$. We use a proper transformation $T(\cdot)$ (according to the camera position relative to the center of the robot) to project the robot's position to the image space, such that $(p_{1_k}, p_{2_k}) = T(p_{x_k}, p_{y_k})$, as shown in \eqref{eq:mu-function}, where $I_T(p_{1_k},p_{2_k})$ is the interpolated pixel value from the output traversability image at pixel location $(p_{1_k},p_{2_k})$.

\begin{equation}
    \mu(p_{x_k},p_{y_k})
    = 
    \begin{cases}
      0 & \text{if $(x_k,y_k) \not\in \mathcal{F}$} \\
      I_T(p_{1_k},p_{2_k}) & \text{otherwise}
    \end{cases}
    \label{eq:mu-function}
\end{equation}

By defining traversability to be zero when the position states $p_x$ and $p_y$ are not part of the field-of-view set $\mathcal{F}$, the model predictive control solution is constrained within the camera's field-of-view, since the kinodynamic model is uncontrollable when the traversability coefficient is zero. The resulting kinodynamic model used by the controller is shown in \eqref{eq:mpc-model}.

\begin{equation}
    \dot{x}(t)
    =
    \begin{bmatrix}
        \dot{p}_x(t) \\
        \dot{p}_y(t) \\
        \dot{\theta}(t)
    \end{bmatrix}
    =
    \begin{bmatrix}
        \mu(p_x(t), p_y(t)) \cdot cos(\theta(t)) & 0 \\
        \mu(p_x(t), p_y(t)) \cdot sin(\theta(t)) & 0  \\
        0 & \nu
    \end{bmatrix}
    \begin{bmatrix}
        v(t) \\
        \omega(t)
    \end{bmatrix}
    \label{eq:mpc-model}
\end{equation}

To design the online reference tracking NMPC, we choose an optimization horizon $N \in \mathbb{N}$ and positive definite matrices $Q$, $R$, and $Q_N$. The following finite horizon optimization formulation is solved to obtain a sequence of control actions that minimizes the cost function
\begin{align} \label{eq:optimization}
    \begin{gathered}
        \min_{x_k, u_k} \sum_{i=k}^{k+N-1} \left\{ ||x_i-x_i^r||^2_Q + ||u_i||^2_R \right\} \\
        + ||x_{k+N}-x_{k+N}^r||^2_{Q_N} + W \cdot F_{\mu}(x_{k:k+N})
    \end{gathered}
\end{align}
The first term minimizes the error between predicted states $x_i$ and the reference state $x^r_i$. The second term minimizes the control action $u_i$ to introduce some damping effect to the control. The third term minimizes the terminal state error. And finally, the fourth term maximizes the clearance based on the traversability around each predicted state. To maximize the clearance around predicted states, we sample points in a uniform distribution around the states and calculate the average for each state. Then we minimize $1-\mu_{avg}$, where $\mu_{avg}$ is the average of traversability for these random points. This is possible because the traversability prediction is in the image space, which enables the sampling of points. $W$ is a weighing cost to fine-tune this term.

\begin{figure}[htp]
    \centering
    \includegraphics[trim={7cm 2cm 7cm 2cm},clip,width=0.92\linewidth]{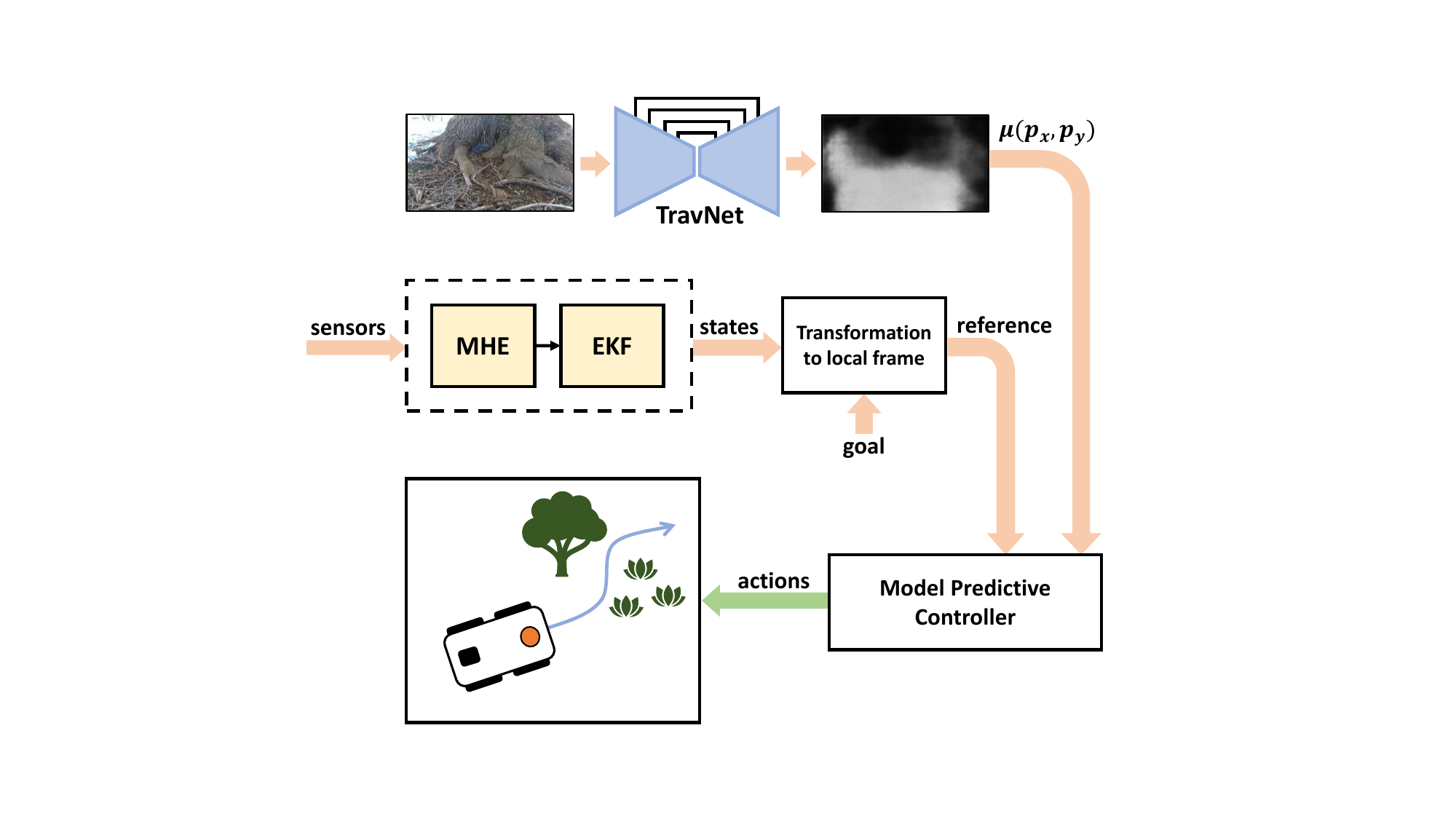}
    \caption{WayFAST modular architecture. Images are used to predict a local traversability map, which is used to generate a cost function for the model predictive control (MPC) block. MPC generates locally optimal goal-oriented trajectories of good traction that avoid obstacles.} 
    \label{fig:mpc-diagram}
    \vspace{-0.1in}
\end{figure}

Let $(u_{i}^\ast)_{i=k}^{k+N-1}$ be the solution for the optimization described in \eqref{eq:optimization}. Then, the first control $u_{k}^\ast$ is applied to the system at time $k$ to drive the robot. Figure \ref{fig:mpc-diagram} shows how TravNet is used in the model predictive control formulation to safely drive the robot in an unstructured environment. To speed up the optimization process on a GPU, we utilize a sampling based MPC approach called MPPI as described in \cite{williams2017model,williams2017information}, where action trajectories are sampled and evaluated to find the minimum cost path within the camera field of view. The controller is parallelized on the GPU and can sample more than 4000 trajectories at a frequency of about 10Hz in real-time.

\section{Experimental Results}

In this section we present the experimental results from different environments. We perform experiments in environments that are similar to the training dataset (forests) but far from the location where we collected the training dataset. We also show generalization experiments in environments that are distinct from the ones present in the training data. We compare our method against baselines to show the effectiveness of our presented method in a variety of scenarios.


\subsection{Traversability Predictions}

\begin{figure}[thpb]
    \begin{subfigure}{.48\linewidth}
        \includegraphics[width=\linewidth]{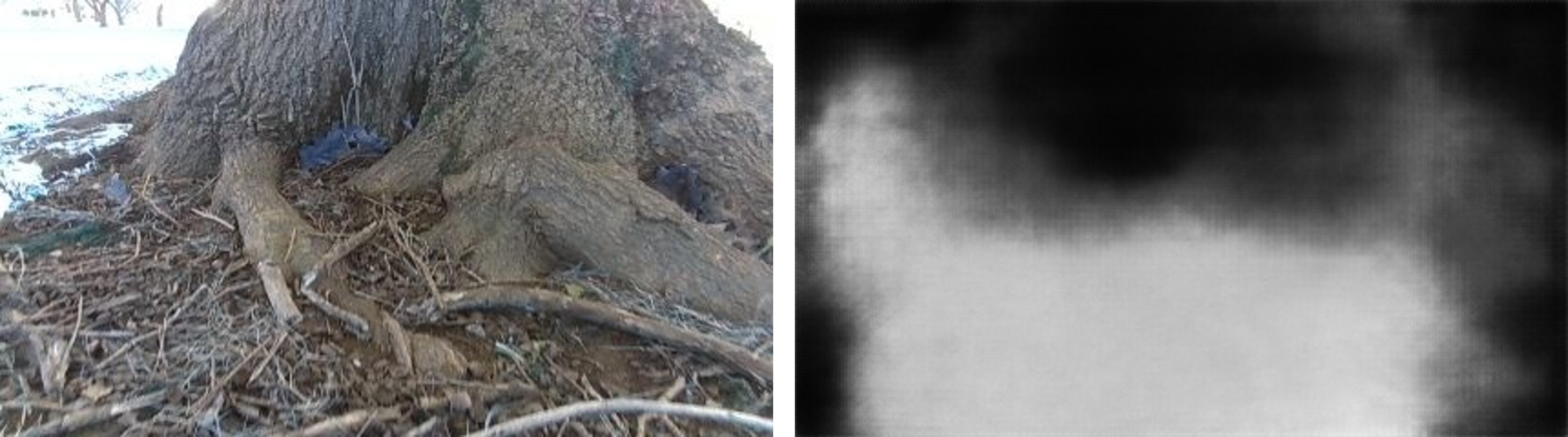}
        \caption{}
        \label{fig:net-outputs-a}
    \end{subfigure}
    \hfill
    \begin{subfigure}{.48\linewidth}
        \includegraphics[width=\linewidth]{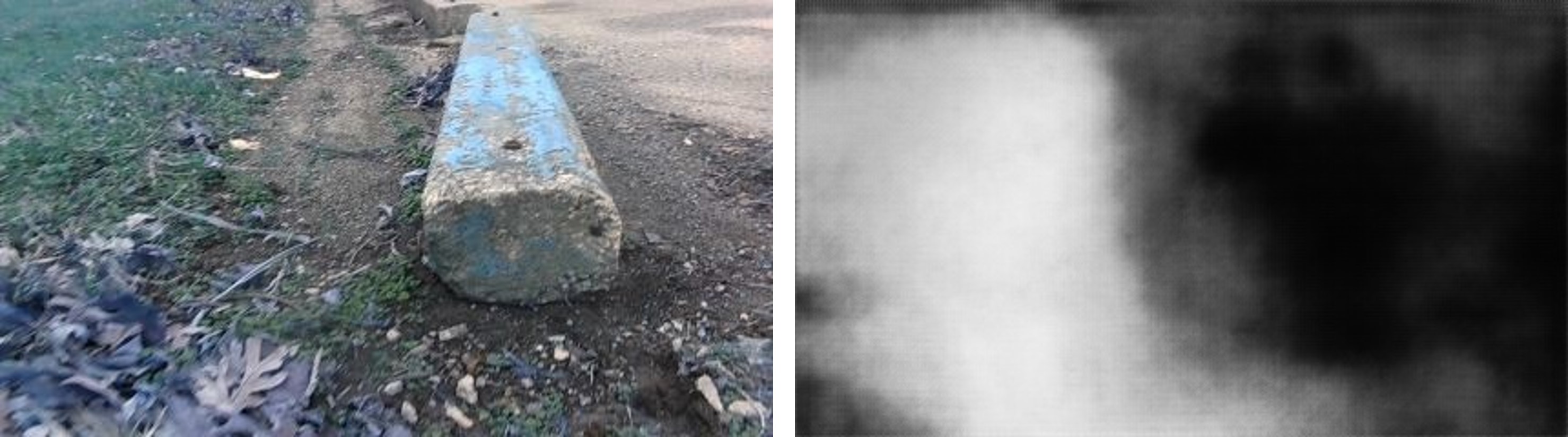}
        \caption{}
        \label{fig:net-outputs-b}
    \end{subfigure}
    
    \begin{subfigure}{.48\linewidth}
        \includegraphics[width=\linewidth]{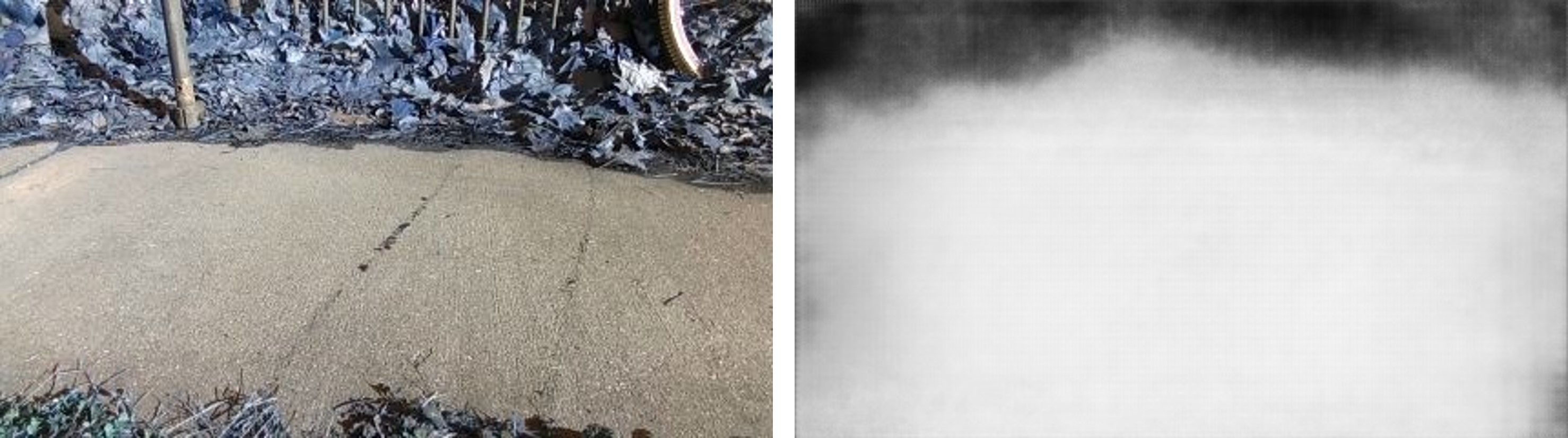}
        \caption{}
        \label{fig:net-outputs-c}
    \end{subfigure}
    \hfill
    \begin{subfigure}{.48\linewidth}
        \includegraphics[width=\linewidth]{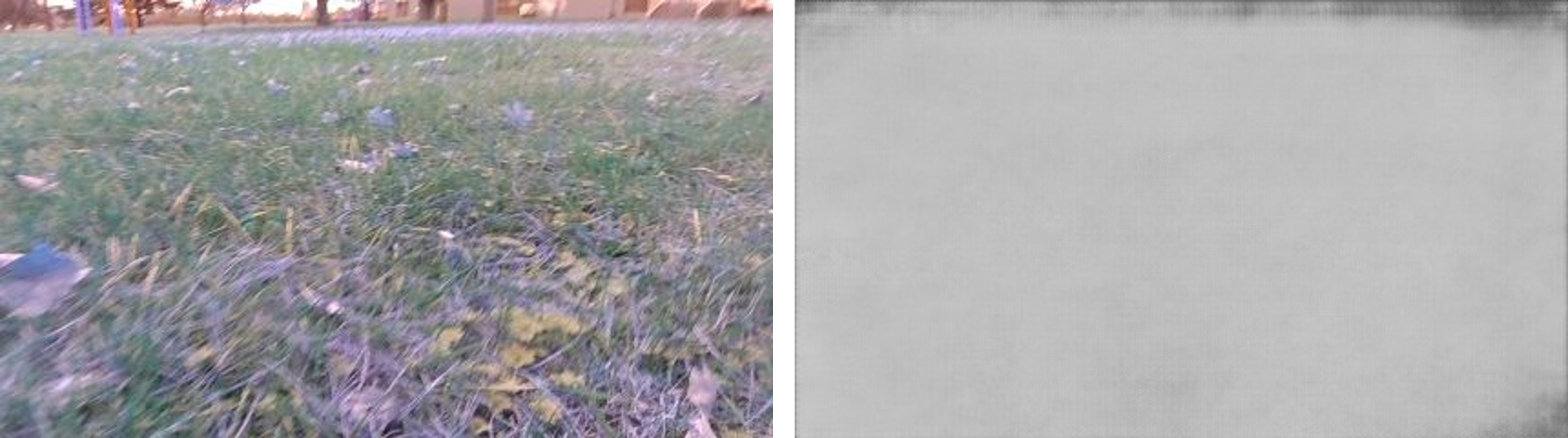}
        \caption{}
        \label{fig:net-outputs-d}
    \end{subfigure}
    \caption{Examples of TravNet inputs in the left images and outputs in the right images. High pixel intensity indicates high traversability value. The examples show \subref{fig:net-outputs-a} a tree trunk, \subref{fig:net-outputs-b} a concrete block, \subref{fig:net-outputs-c} a bicycle rack with a concrete path at the front, and \subref{fig:net-outputs-d} flat grass terrain.}
    \label{fig:net-outputs}
\end{figure}

Using RGB and depth modalities as input, we found that TravNet is successful in predicting traversability for a variety of obstacles. Figure \ref{fig:net-outputs} shows a variety of terrain examples in which TravNet was used to predict the traversability coefficients. 

\begin{figure}[thpb]
    \centering
    \includegraphics[trim={0cm 5cm 0cm 6cm}, clip, width=0.96\linewidth]{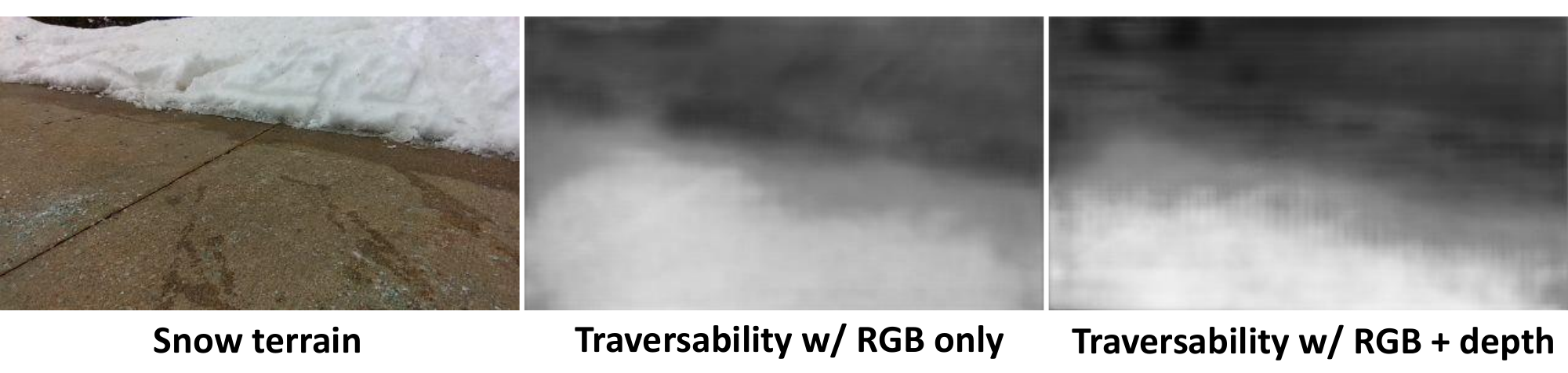}
    \caption{The left image shows snow terrain that would not be identified using only geometric methods. The center image shows predicted traversability using only RGB. The right image shows the prediction using RGB and depth images.}
    \label{fig:snow-terrain}
    \vspace{-0.1in}
\end{figure}

In addition, we show that TravNet is able to predict semantic information that would not be possible with only geometric information. Fig. \ref{fig:snow-terrain} shows an example of snow traversability prediction. Such prediction would be problematic for a sensor that can only capture geometric measurements, since the height and texture are similar to regular terrain. However, our model is able to predict that snow is less traversable than grass, and we show that such semantic understanding is conserved with the addition of a geometric modality (depth camera).

Although the use of TravNet without the depth modality is possible, we noticed the depth measurement helps with generalization to unseen environments. This improvement is noticeable on environments with higher numbers of geometrical obstacles, such as a wall with different textures and colors from the ones seen in the training dataset.

\subsection{Comparison with Baseline Approaches}

We compare our approach (WayFAST) with BADGR, a state-of-the-art learning based approach presented in \cite{kahn2021badgr}. In addition, we also compare against 2 other baselines - 1. a naive method that blindly follows the goal (Blind Pursuit) using the same kinodynamic model and MPPI used in WayFAST but disregarding the traversability map, 2. a LiDAR-based algorithm that generates a traversability map according to geometric obstacles. Both Blind Pursuit and the LiDAR-based method use our same control framework, with the only modification being the perception system. We specify a point goal location and initially position the robot to face the opposite direction for all experiments.

\begin{figure}[htp]
    \centering
    \includegraphics[width=0.95\linewidth]{{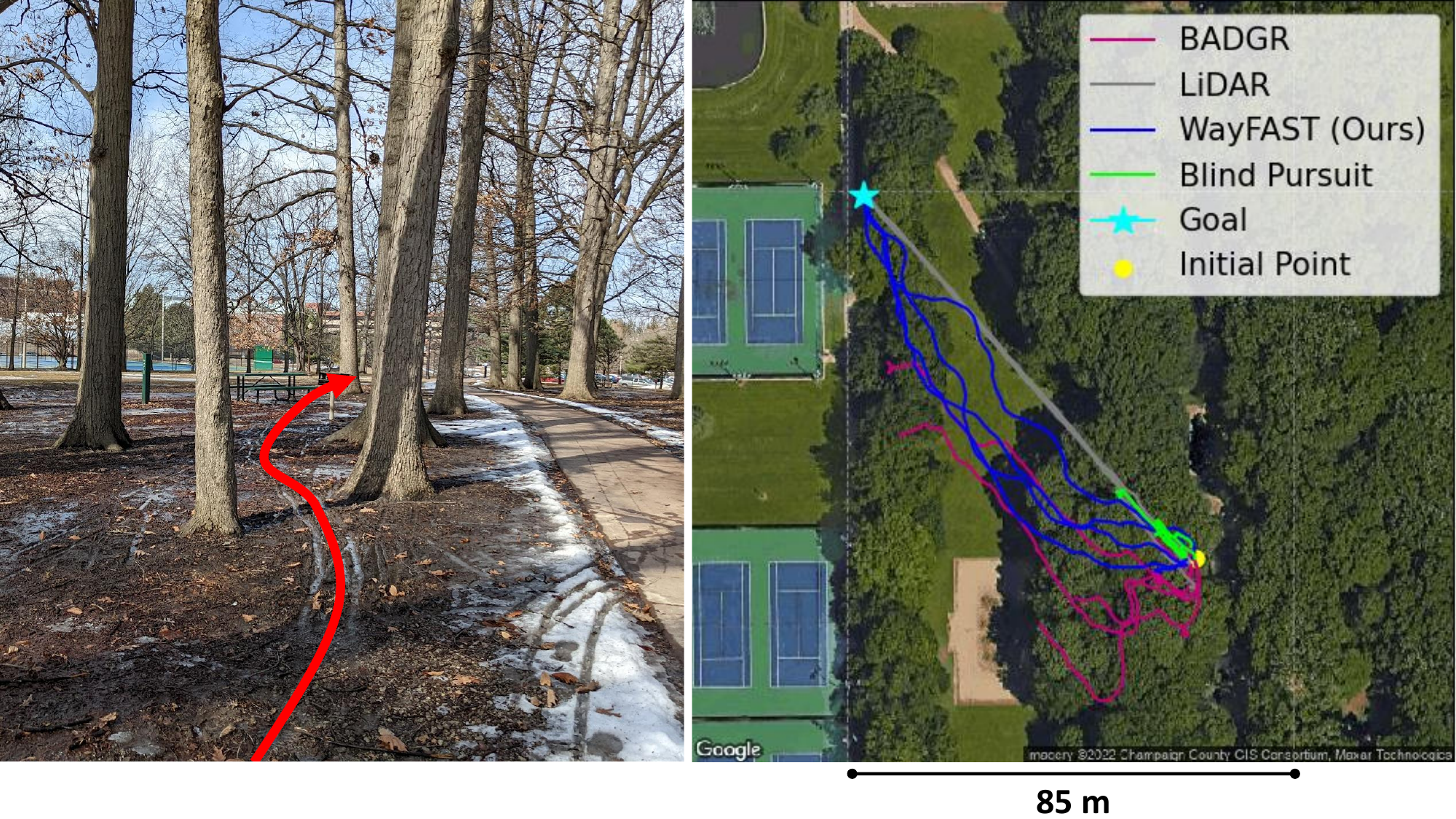}}
    \caption{Comparison between WayFAST (ours) and baselines. The left image shows the environment and a path the robot would need to traverse to reach the goal. Baselines include BADGR \cite{kahn2021badgr}, a LiDAR-based method, and Blind Pursuit without any use of traversability.}
    \label{fig:comparison-badgr}
\end{figure}

Table \ref{table:forest} and Fig. \ref{fig:comparison-badgr} show our results in a forest-like environment, where WayFAST was able to reach the goal as many times as the LiDAR baseline. It is important to note that LiDAR has a 270 degree field-of-view and a longer detection range, while WayFAST only uses a camera with a 69 degree field-of-view that points slightly downwards. This allows the LiDAR baseline to provide a larger set of path solutions. Since the LiDAR baseline uses all of the components from WayFAST (i.e., the kinodynamic model, model predictive controller, and state estimator), we expect that similar performance could be obtained for WayFAST with the same range of predictions. The blind pursuit baseline always crashed into an obstacle or got stuck in untraversable areas. BADGR was able to navigate for a long period but never reached the goal. It could be because BADGR learns not only the traversability representation but also the robot's dynamic model and our dataset was not rich enough for the neural network to properly learn the necessary dynamics. In addition, WayFAST uses depth and skip connections, which can help in the learning process for small datasets.

\begin{table}[th]
\centering
\begin{tabular}{|c|c|c|c|}
\hline
\multicolumn{1}{|l|}{} & \textbf{Total tries} & \textbf{Successful runs} & \textbf{Success rate} \\ \hline
\textbf{WayFAST}       & 5                    & 4                        & 80\%                  \\ \hline
\textbf{BADGR}         & 5                    & 0                        & 0\%                   \\ \hline
\textbf{LiDAR-based}   & 5                    & 4                        & 80\%                  \\ \hline
\textbf{Blind Pursuit} & 5                    & 0                        & 0\%                   \\ \hline
\end{tabular}
\caption{Summary of experiments in a forest-like environment, where our approach (WayFAST) is comparable to a LiDAR baseline while outperforming the other two baselines.}
\label{table:forest}
\vspace{-0.1in}
\end{table}

\subsection{Comparison with Classical Geometric Navigation}

\begin{figure}[htp]
    \centering
    \includegraphics[trim={1.2cm 0cm 1.2cm 0cm}, clip, width=0.96\linewidth]{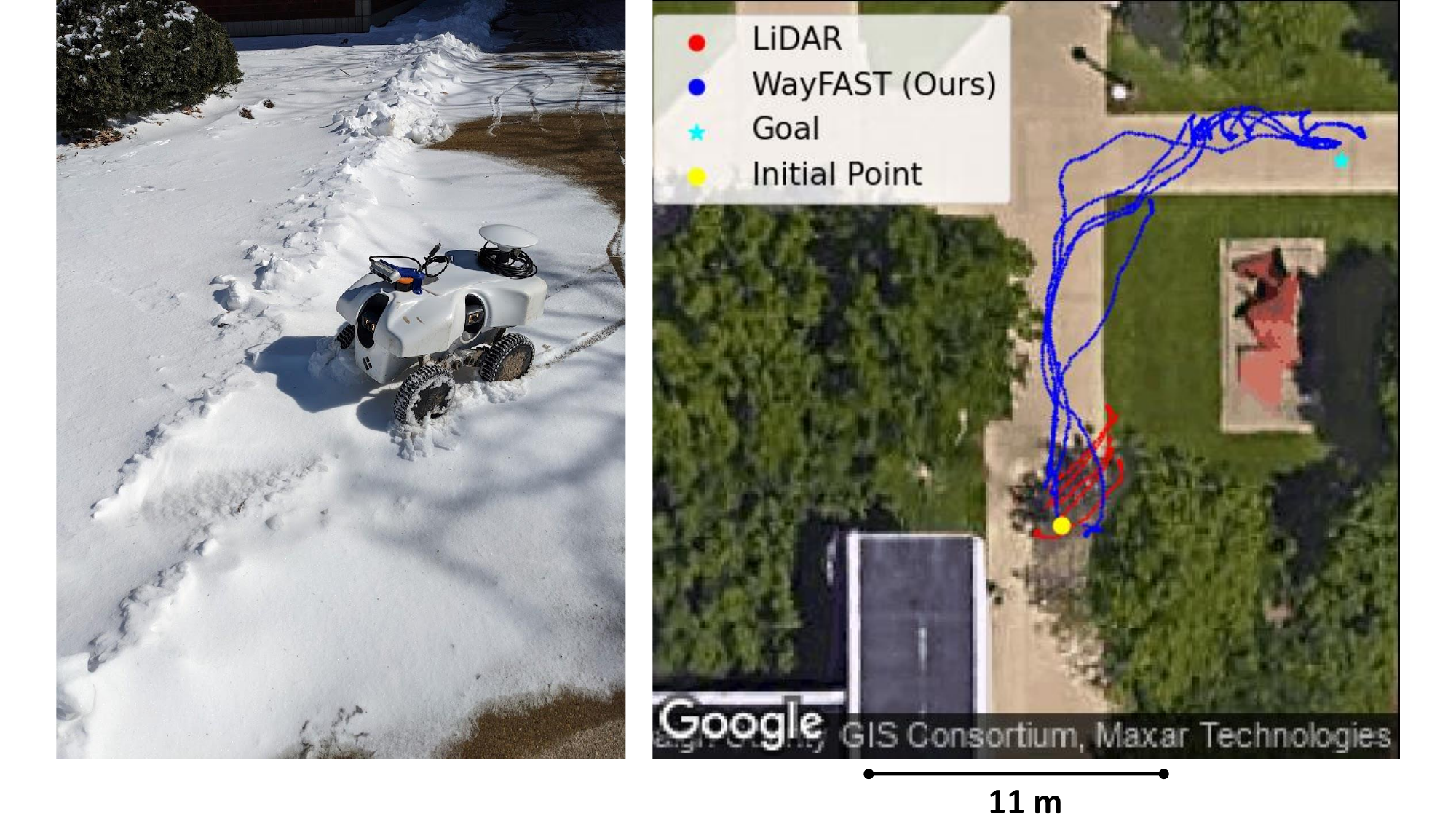}
    \caption{Comparison between WayFAST and LiDAR navigation on snow terrain (snow not captured by the overhead satellite image). The left image shows how navigation fails on snow when using the LiDAR-based method.}
    \label{fig:comparison-lidar}
    \vspace{-0.1in}
\end{figure}

In order to show that geometric information is not enough for unstructured outdoor environments, we performed additional experiments comparing our proposed approach to LiDAR-based navigation. We tested both methods in areas with snow, and as shown in Fig. \ref{fig:comparison-lidar}, LiDAR was never able to reach the goal since it always got stuck in snow. This experiment shows that WayFAST was able to perceive the snow as an untraversable area and therefore safely avoid it to reach the goal. As shown in Fig. \ref{fig:comparison-lidar}, in four of the five experiments, WayFAST safely reached the goal. In the other experiment, WayFAST reached a location close to the goal but got stuck along the way. The LiDAR-based geometric method was not able to perceive the snow and therefore failed to navigate early in its trajectories.

\subsection{Generalization for Unseen Environments}

\begin{figure}[htp]
    \centering
    \includegraphics[trim={0cm 1.2cm 0cm 2cm}, clip, width=0.96\linewidth]{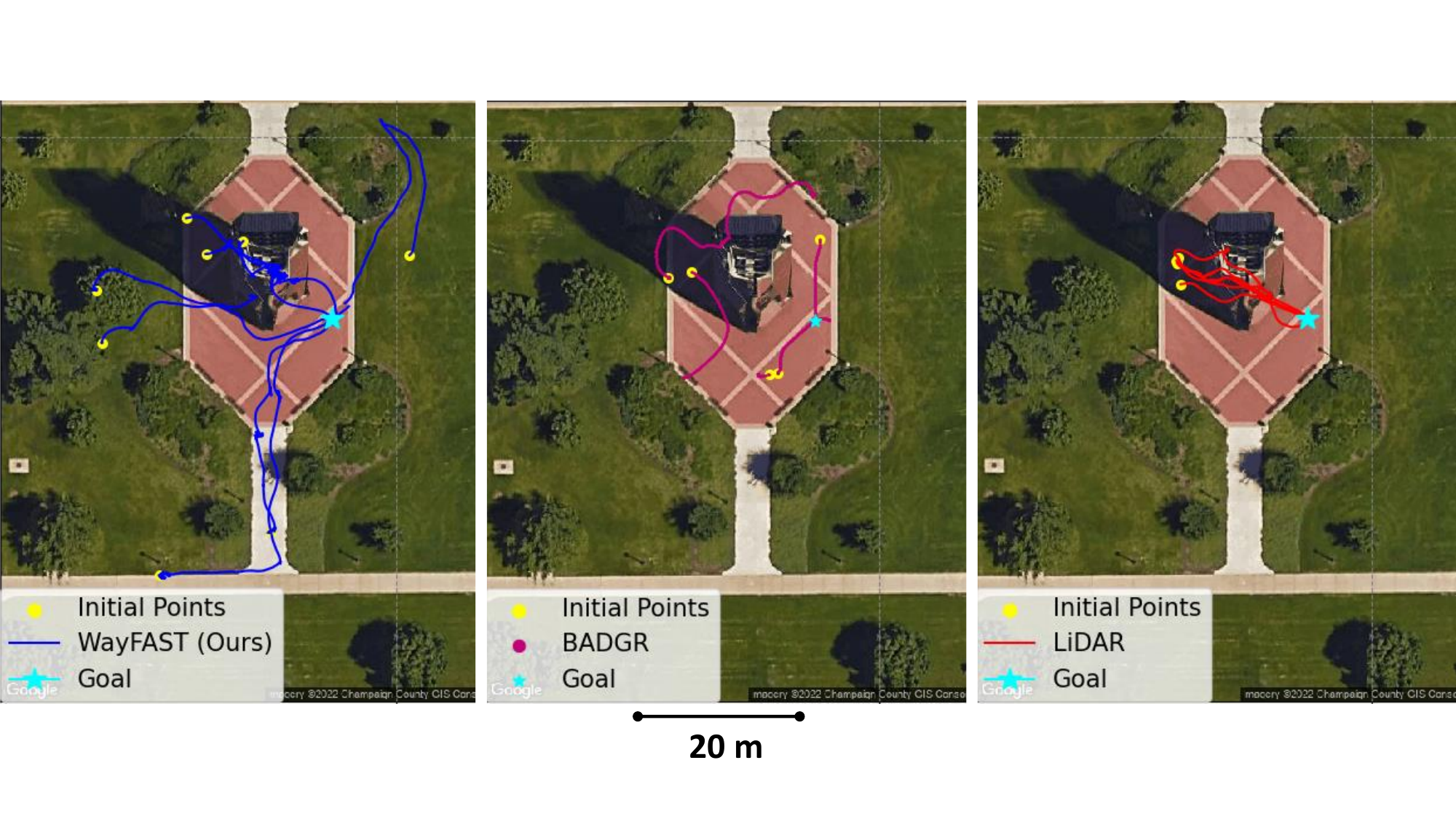}
    \caption{Left image shows WayFAST starting from eight different points with same goal. The longest path presented in the left image is due to a snow patch between the starting point and the building (not captured in satellite image). Center image presents BADGR \cite{kahn2021badgr} and the  right image shows LiDAR-based navigation starting from five different locations with same goal.}
    \label{fig:generalization}
    \vspace{-0.1in}
\end{figure}

We tested our method for generalization on a very different environment from the one in the dataset. We chose an urban environment with brick walls and narrow spaces, which can be mostly represented as simple geometric obstacles. Figure \ref{fig:generalization} shows that our method can generalize well to such environments, despite being trained on a dataset that does not include representative images. Our hypothesis is that depth as an additional input modality assists in generalization to avoid geometric obstacles in new environments different from the training set. To test this hypothesis, we evaluated our method in the same environment as shown in Fig. \ref{fig:generalization}, with and without the use of depth images. We repeated the test five times for each case. Our results show RGB with depth images to be more robust than without depth, reaching the goal successfully in all repetitions. Our method reached the goal three out of five times when using only RGB images as input.

\subsection{Navigation with Only RGB Data}

\begin{figure}[htp]
    \centering
    \includegraphics[trim={1.4cm 0cm 1.4cm 0cm}, clip, width=0.95\linewidth]{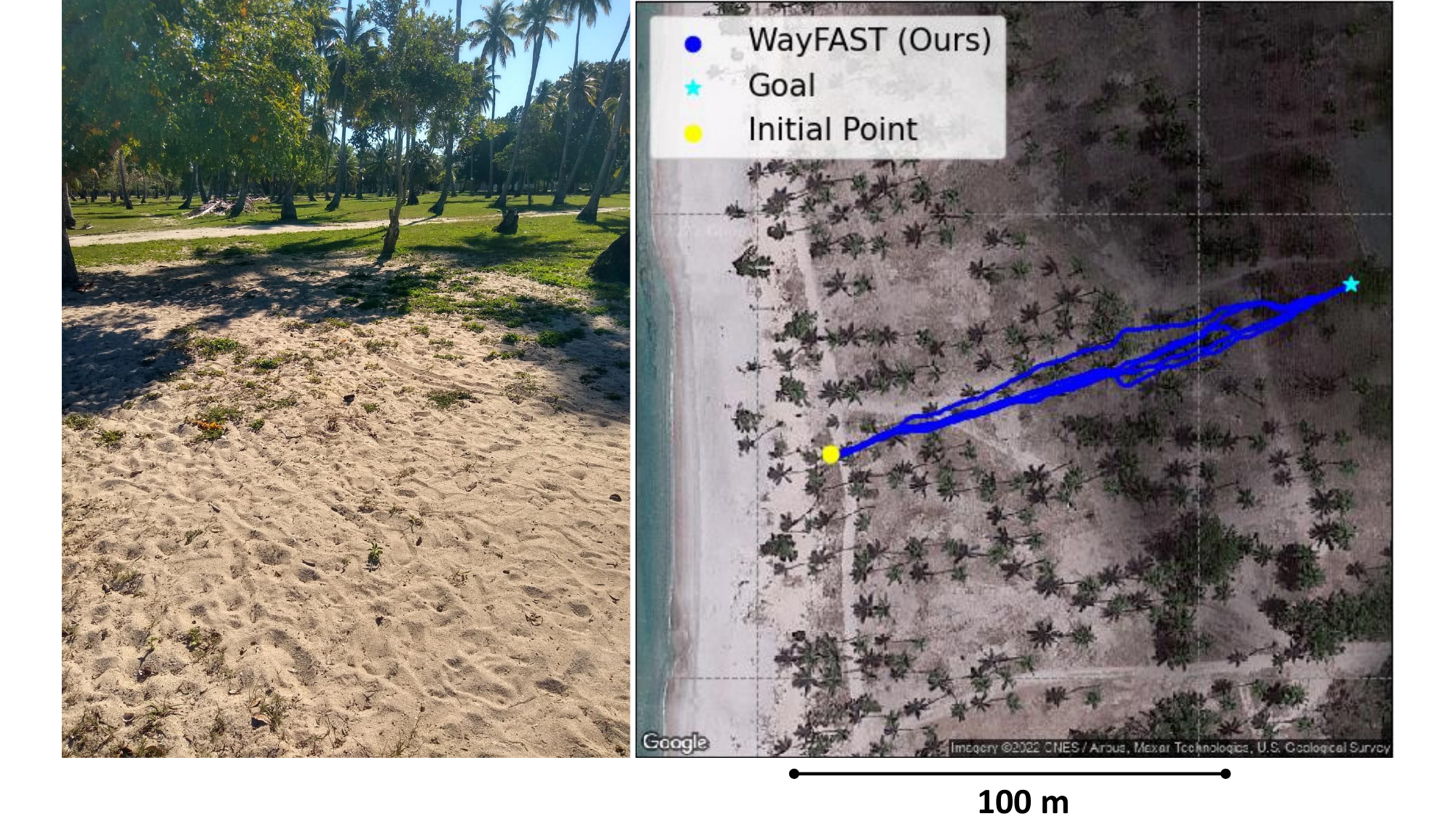}
    \caption{Experimental results in coconut plantation without depth information. Left image shows the terrain.}
    \label{fig:coconut-experiment}
\end{figure}

We tested our method with only RGB data as input for TravNet removing the branch responsible for encoding depth information. This test was performed in Puerto Rico in a different scenario from the training dataset. Figure \ref{fig:coconut-experiment} shows that WayFAST was able to successfully navigate on a challenging outdoor environment even with only RGB input. We performed five runs starting from the same location and with the same goal. The robot was able to complete all five runs without getting stuck or colliding into obstacles.

\begin{figure}[htp]
    \centering
    \includegraphics[trim={3cm 0cm 3cm 0cm}, clip, width=0.95\linewidth]{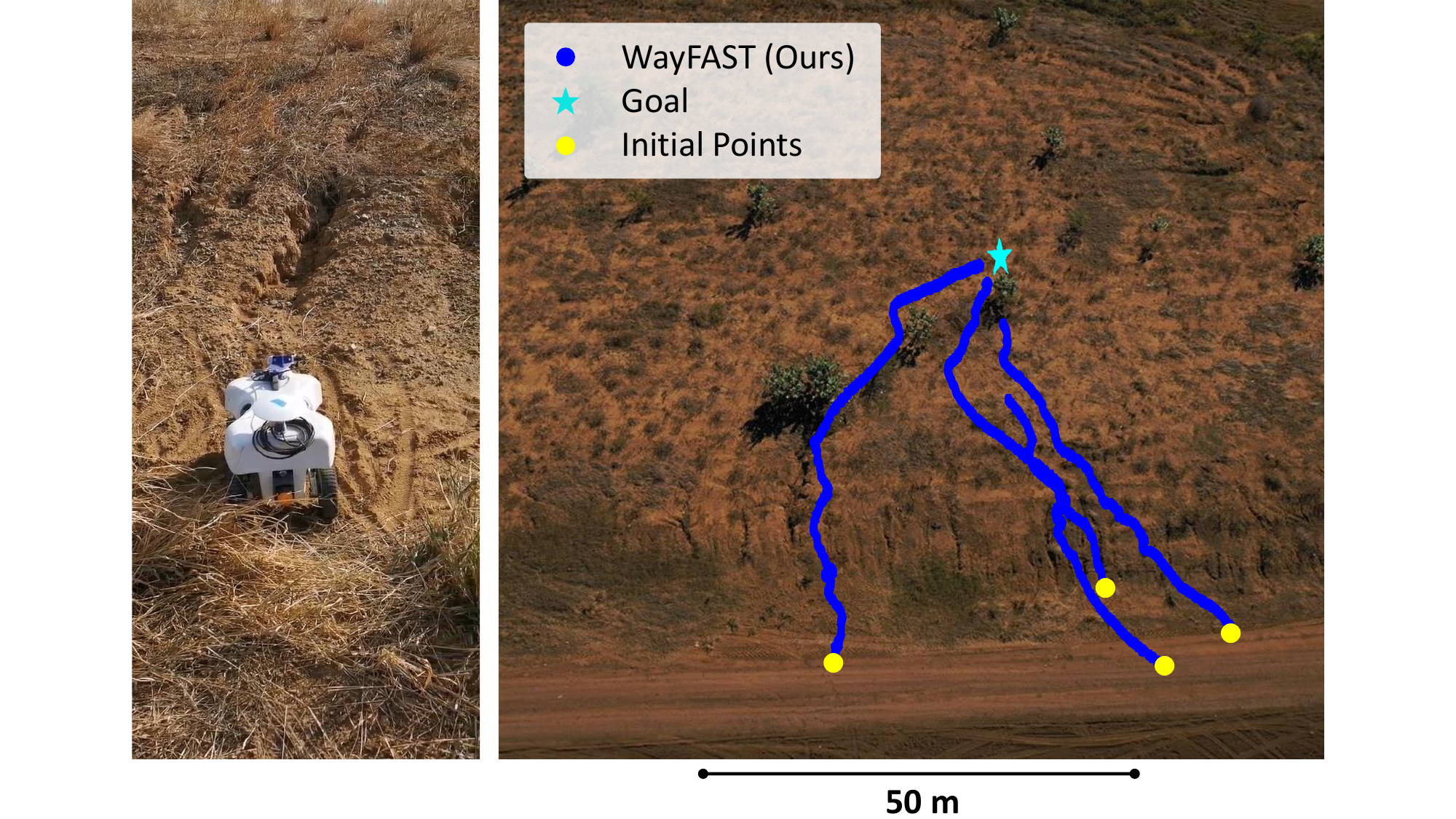}
    \caption{Experimental result on uneven terrain. The left image shows example depressions and unevenness in the terrain. The right image shows paths for the four runs.}
    \label{fig:uneven-terrain}
    \vspace{-0.2in}
\end{figure}

Figure \ref{fig:uneven-terrain} shows a test in uneven terrain where only RGB was used as input. Also the environment was different from the one present in the dataset. We started the robot at different locations and evaluated if the robot could avoid depressions and obstacles. Out of four runs, the robot arrived close to the goal three times and got stuck in the soil in one run.

\subsection{Limitations and Failure Modes}

From our experiments, we found that a major limitation was the camera's field-of-view. Since the camera used in the experiments has a narrow field-of-view, and the RealSense depth measurements have only a short 5 meter range, the navigation algorithm was not able to plan for long horizons and explore a large variety of paths. The model predictive controller is constrained to find paths in the camera's field-of-view, and therefore, find a minimal cost for that space. With a reduced space for minimization, the robot was not able to avoid large objects or plan around large untraversable areas.

In addition, snow is an example of a situation that may be traversable or untraversable, depending on the softness and depth, which can change with weather. However, in both cases, the snow often has the same visual appearance. We found that even after adding snow data to the dataset, the traversability for snow could vary depending on location, and other sensor modalities may therefore be necessary to safely detect when snow is traversable or untraversable.

\section{Conclusion}

We presented a self-supervised traversability prediction system for autonomous navigation of wheeled robots in unstructured outdoor environments. Our approach uses a modular architecture that combines traversability prediction using a convolutional neural network that fuses RGB and depth inputs and known kinodynamic model to autonomously navigate in a variety of outdoor environments. We validated our method in different terrains and show better navigation performance than other state-of-the-art approaches. In addition, we demonstrate that our method can overcome limitations present in classical approaches due to the lack of semantic understanding, such as in the avoiding snow. We hope our work leads to further research in this area thereby enabling field robots to navigate without predefined waypoints in unstructured outdoor environments.

\addtolength{\textheight}{-1cm}   

\bibliographystyle{IEEEtran}
\bibliography{references}

\end{document}